%% file: colm2026_conference.tex
\documentclass{article} % For LaTeX2e
\usepackage[preprint]{colm2026_conference}
\usepackage[T1]{fontenc}

\usepackage{microtype}
\usepackage{hyperref}
\usepackage{url}
\usepackage{booktabs}

\usepackage[most]{tcolorbox}
\usepackage{xcolor}
\usepackage{enumitem}
\usepackage{booktabs}
\usepackage{tabularx}
\usepackage{array}
\usepackage{inconsolata}
\usepackage[table]{xcolor}
\usepackage{longtable,array,multirow}
\usepackage[table]{xcolor}
\usepackage[normalem]{ulem}
\usepackage[most]{tcolorbox}
\usepackage{tikz}
\usepackage{array}
\usepackage{pifont}
\usepackage{booktabs}
\usepackage{threeparttable}
\newcommand{\cmark}{\ding{51}}

\newtcolorbox{judgecheck}[2][]{%
  enhanced,
  colback=white,
  colframe=#2!80!black,
  boxrule=0.9pt,
  arc=7pt,
  left=6pt,
  right=6pt,
  top=5pt,
  bottom=5pt,
  drop shadow={black!12},
  #1
}

\usepackage{tikz}
\usetikzlibrary{positioning,arrows.meta}
\usepackage{booktabs,longtable,array}
\newcolumntype{P}[1]{>{\raggedright\arraybackslash}p{#1}}

\newcolumntype{L}[1]{>{\raggedright\arraybackslash}p{#1}}
\newcolumntype{Y}{>{\raggedright\arraybackslash}X}

\newcommand{\srcspan}[1]{\colorbox{green!20}{#1}}

\newcommand{\srcHL}[1]{\colorbox{green!20}{\strut #1}}
\newcommand{\outHL}[1]{\colorbox{blue!18}{\strut #1}}
\newcommand{\trHL}[1]{\colorbox{orange!25}{\strut #1}}
\newcommand{\tgtspan}[1]{\colorbox{blue!18}{#1}}   % use your purple if preferred
\tcbset{
  auditbox/.style={
    colback=white,
    colframe=black!40,
    boxrule=0.6pt,
    arc=2pt,
    left=5pt,right=5pt,top=4pt,bottom=4pt
  }
}
\usepackage{xcolor}
\usepackage{pifont}
\usepackage{tikz}
\usetikzlibrary{positioning,arrows.meta,calc,fit}
\usepackage{pifont}

\usepackage{listings}
\newcommand{\xmark}{\textcolor{red}{\ding{55}}}

\usepackage{lineno}

\definecolor{darkblue}{rgb}{0, 0, 0.5}
\hypersetup{colorlinks=true, citecolor=darkblue, linkcolor=darkblue, urlcolor=darkblue}
\usepackage{xcolor}
\usepackage{soul}
\definecolor{errhighlight}{RGB}{253, 216, 206}  % light coral
\newcommand{\errtxt}[1]{\sethlcolor{errhighlight}\hl{#1}}

\title{Should We be Pedantic About Reasoning Errors in Machine Translation?}
\author{Calvin Bao, Marine Carpuat \\
Department of Computer Science \\
University of Maryland \\
College Park, MD 20742, USA \\
\texttt{csbao@umd.edu}
}
\usepackage{listings}
\usepackage{algorithm}
\usepackage{algpseudocode}

\input{comments}
\begin{document}

\ifcolmsubmission
\linenumbers
\fi

\maketitle

\begin{abstract}

% Address faithfulness question more directly with general perturbation techniques
% Reasoning doesnt help MT, we dont knwo why
% paper establishes there are reasoning errors in the traces
% we attempt to intervene with simple tweaks guided by the detected issues, but it doesnt work that well
% using a reasoning trace from a larger model, would this be a "stronger" intervention?
% story is more -> with presumably higher quality traces, we can get better translation quality, as an intervention. this is just analytical / diagnosis. can we improve quality by making the reasoning better?
% motivates future work on how to get models to reason better in this setting? something that uses the reasoning generated from other approaches?
% Large reasoning models produce reasoning traces before translating, yet whether these traces are correct, whether the model conditions on them, and whether correcting their errors improves translation remain open questions. We investigate all three across 
% % two reasoning models (Qwen3-8B, 
% Ministral-3-8B) and seven language pairs spanning high-resource (English $\to$ \{Spanish, French, German, 
% Chinese, Japanese\}) and lower-resource 
% (English $\to$ \{Urdu, Cantonese\}) directions.

Across multiple language pairings (English $\to$ \{Spanish, French, German, Mandarin, Japanese, Urdu, Cantonese\}), we find reasoning errors in translation. To quantify how often these reasoning errors occur, we leverage an automated annotation protocol for reasoning evaluation wherein the goal is to detect if a reasoning step is any of three error categories: (1) source sentence-misaligned, (2) model hypothesis-misaligned, or (3) reasoning trace-misaligned. We probe the reasoning model with perturbed traces correcting for these identified reasoning errors using an array of weak-to-strong interventions: hedging, removal, re-reasoning after removal, hindsight, and oracle interventions. Experimenting with interventions on the reasoning traces suggests that small corrections to the reasoning have little impact on translation quality, but stronger interventions yield the highest resolution rates, despite translation quality gains being mixed.  We find ultimately that reasoning errors in MT can be identified with high precision in Urdu but lower precision in Spanish, but that removing these reasoning errors does not resolve the initial errors significantly, suggesting limited reasoning faithfulness for machine translation.

\end{abstract}

\input{sections/intro-2}

\input{sections/background-2}

\input{sections/framework}

\input{sections/intervention}

\input{sections/intervention2}

\input{sections/conclusion}

\section*{Acknowledgments}
We thank the members of the UMD CLIP lab for their valuable advice and feedback on earlier iterations of this work, especially Sarah Wiegreffe, Dayeon Ki, and Hyojung Han.
% Use unnumbered first level headings for the acknowledgments. All
\section*{Limitations}
\paragraph{Scope of models and languages.}
Our experiments are conducted on two open-weight reasoning models (Qwen3-8B and Ministral-3-8B), both at the 8B parameter scale. It is unclear whether the patterns we observe hold for 
other reasoning models. We test 
seven language pairs, all with English as the source. Translation from non-English sources, or between two 
non-English languages, may surface qualitatively different reasoning errors.

\paragraph{Limited human validation coverage.}
We conduct bilingual human validation for only two 
of our seven language pairs (English$\to$Urdu and 
English$\to$Spanish), which were chosen to contrast a 
lower-resource and a high-resource setting. The remaining five pairs lack direct human validation of the judge's precision, for cost reasons. Given that we observe different precision rates between Urdu (93.1\%) and Spanish (50.0\%), it is probable that 
precision varies across the unvalidated pairs as well, and we cannot determine in which direction without additional annotation.

% \paragraph{Scope of claims.}
% Our experiments are conducted on two open-weight models (Qwen3-8B and Ministral-3-8B) across five language pairs, all 
% involving English as the source language and high-resource target languages. It is presumable that these findings will not generalize to low-resource languages, and that low-resource languages may have qualitatively different reasoning errors that make translation difficult in these settings. As is the case for much empirical work, the error rates 
% and intervention outcomes we report are specific to the models, 
% data, and experimental setup described in this paper.

\section*{Ethics Statement}
% \cb{Stub for LLM-as-a-judge precision against human annotators, and how closed-source models exposed by private APIs are not a reliable way to build our foundation over}

\paragraph{Intended use.}
The work presented here aims to characterize where and how reasoning traces may fail in machine translation, 
not as a proposal for improving translations in production. Our intervention experiments show that 
targeted trace edits can resolve some detected errors but do not reliably improve overall translation quality. 
We do not recommend the interventions described here as a post-editing pipeline without further quality 
assurance, as they may introduce new errors not present in the original translation.

\paragraph{LLM disclosure.}
LLMs were used in several different ways in the preparation of this research. For paper writing, Anthropic Claude was used for drafting, refining language (grammar and style), and creating LaTeX tables and figures. For coding, Claude was used to refine scripts for model generation and data analysis. Finally, for research ideation, \cite{elicit} was used to guide and help summarize literature review. All scientific claims, ultimate experimental design decisions, and interpretations are the authors' own.

\bibliography{colm2026_conference}

\bibliographystyle{colm2026_conference}

\appendix
\section{Appendix}
\subsection{Detecting reasoning errors}

As discussed in the main paper, we detect reasoning errors in three categories. We show prompt templates.

\begin{lstlisting}
You are a bilingual auditor for machine-translation reasoning traces.

You will analyze a SOURCE sentence, the model's TRACE (reasoning while translating), and the OUTPUT (final translation).

Your task is to detect reasoning errors in three categories:
1. INPUT_TRACE: Trace statements not supported by SOURCE, or proposing incorrect translation semantics (e.g., hallucinated facts, wrong word meanings)
2. TRACE_OUTPUT: Trace decisions that don't match the OUTPUT (e.g., trace says "X" but output has "Y")
3. TRACE_INTERNAL: Contradictions, circular reasoning, or incoherent statements within the trace itself

IMPORTANT RULES:
- The trace will be sentence-tokenized. Reference issues by sentence index (0-indexed).
- All quotes must be EXACT substrings (copy-paste) from the provided text.
- Be strict but fair - minor rephrasing or stylistic choices are not errors.

Output ONLY valid JSON matching this schema:
{
  "has_issues": bool,
  "summary": str,  // One sentence summary of trace quality
  "issues": [
    {
      "category": "INPUT_TRACE" | "TRACE_OUTPUT" | "TRACE_INTERNAL",
      "trace_sentence_idx": int,
      "trace_quote": str,  // Exact substring from trace
      "source_quote": str | null,  // Relevant source quote if applicable
      "output_quote": str | null,  // Relevant output quote if applicable
      "rationale": str  // 1-2 sentence explanation
    }
  ]
}

SOURCE:
{source}

TRACE (sentence-indexed):
[0] {trace_sentence_0}
[1] {trace_sentence_1}
...

OUTPUT:
{output}
\end{lstlisting}

% \paragraph{LLM Judge User Prompt Template.}
% \begin{lstlisting}

% Analyze the trace for issues. Return JSON only.
% \end{lstlisting}

% \subsection{Examples grounded by resolution.}
% \input{tables/replay_examples}

\subsection{Bilingual annotation interface}
See \autoref{fig:bilingual_annotation_protocol} for a flow of the annotation protocol.

\input{figures/human_eval}
\subsection{Intervention Implementation}\label{app:replay-shared}
% Optional (uncomment what you use):
% \usepackage{listings}
% \lstset{basicstyle=\ttfamily\footnotesize,breaklines=true,columns=fullflexible}
% \usepackage{algorithm}
% \usepackage{algpseudocode}

% \subsection{Shared components for trace replay}

\paragraph{System message (default).}
\texttt{You are a careful machine translation assistant.}

\paragraph{Task instruction (default).}
If source and target language codes are known, the default is:
\texttt{Translate the following $\langle\textit{source language}\rangle$ text into $\langle\textit{target language}\rangle$. Return only the translation.}
(with fallbacks when only the target or neither is known).

\paragraph{Follow-trace instruction (default).}
\texttt{Use the reasoning trace when deciding on the translation.}

\paragraph{User message template.}
Interventions that replay with an (possibly edited) trace all wrap the model input in this pattern (then passed through the model chat template if available):

\begin{lstlisting}
{task_instruction}

Source:
{source}

Reasoning trace:
{edited_trace}
{optional_additional_notes_block}

{follow_trace_note}
Return only the final translation.
\end{lstlisting}

If \texttt{extra\_notes} are supplied (Oracle methods), the block is:

\begin{lstlisting}

Additional notes:
- {note_1}
- {note_2}
...
\end{lstlisting}

Replay generation uses greedy decoding (\texttt{do\_sample=False}, beam width~1). The trace is \emph{not} placed in native ``thinking'' channels for these replays; only the re-reasoning intervention forces a continuation inside the model's reasoning format (see below).

\subsubsection{Localizing the edited span}
\label{app:span-locate}

For hedging, removal, and re-reasoning, we find a character span for the issue in the original trace:

\begin{algorithmic}[1]
\If{issue has non-empty \texttt{trace\_quote}}
    \State find \texttt{trace\_quote} in \texttt{trace} (fuzzy / normalized match)
\Else
    \State use sentence boundaries and \texttt{trace\_sentence\_idx}
\EndIf
\State expand to the \textbf{minimal whole sentence} that contains that span (\texttt{locate\_issue\_edit\_span}), so edits remove or hedge full sentences when possible
\end{algorithmic}

If no span can be located, hedging, removal, and re-reasoning are skipped for that issue.

\subsubsection{Hedging}
\label{app:hedging}

Let $s$ be the sentence text at the edit span. If $s$ is non-empty and does not already begin (case-insensitive) with \texttt{maybe}, \texttt{possibly}, \texttt{perhaps}, or \texttt{it may be}, replace $s$ by:
\begin{quote}\ttfamily\small
Possibly, but this should be verified against the source: $\langle s\rangle$
\end{quote}
Otherwise $s$ is unchanged. The modified trace is replayed with the shared template (Section~\ref{app:replay-shared}).

\subsubsection{Removal}
\label{app:removal}

The sentence-level span from Section~\ref{app:span-locate} is deleted from the trace; consecutive blank lines are collapsed. The shortened trace is replayed with the same user template as the baseline (no extra notes).

\subsubsection{Re-reasoning after removal}
\label{app:rereason}

Let $t$ be the original trace and $[i,j)$ the edit span. The \emph{prefix} is $p = t_{:i}$ (text strictly before the removed span), with blank lines collapsed.

The user-visible instruction (before chat templating) is:

\begin{lstlisting}
{task_instruction}

Source:
{source}

A problematic reasoning step was removed here. Reconsider the source carefully from this point onward and do not rely on the removed unsupported step. Target issue: {issue.rationale}
You MUST continue reasoning internally from the provided starting point, but return only the final translation.
\end{lstlisting}

% For Qwen3-family models, \texttt{/think} is appended to the user content to enable native reasoning. The formatted prompt is then extended by appending an opening thinking block followed by the prefix trace, so the model continues internal reasoning from $p$:
% \begin{quote}\ttfamily\small
% \ldots\texttt{<redacted\_thinking>}\\

% \end{quote}

\subsubsection{Hindsight (reference-guided trace, then replay)}
\label{app:hindsight}

Hindsight uses the same replay model with native reasoning enabled.

\paragraph{Step 1 -- synthesize a trace toward the reference.}
Prompt:

\begin{lstlisting}
{task_instruction}

Source: {source}
Reference translation: {reference}

Think step-by-step about how to translate the source to match the reference.
Analyze key phrases, idioms, and grammatical structures.
Then produce the final translation.
\end{lstlisting}

% The model generates with \texttt{enable\_native\_reasoning=True} (Qwen3: \texttt{/think} suffix in the templated user message). The \emph{reasoning-only} substring is extracted from the decoded continuation: for Mistral/Ministral, the text inside \texttt{[THINK]...\texttt{[/THINK]}}; for Qwen3, content inside \texttt{<redacted\_thinking>...\texttt{</redacted\_thinking>}} (with fallbacks if tags are missing).

\paragraph{Step 2 -- replay.}
The extracted trace $t'$ is substituted into \texttt{Reasoning trace:} in the shared replay template (Section~\ref{app:replay-shared}) with \emph{no} extra notes. One replay is stored per record (not per issue); fix-judging, if enabled, is still evaluated per targeted issue against that single output.

\subsubsection{Direct oracles (\texorpdfstring{Oracle-$1$}{Oracle-1} and \texorpdfstring{Oracle-$K$}{Oracle-K})}
\label{app:oracle}

Oracle interventions keep the \textbf{original} model trace and add natural-language hints derived from the judge/issue object (not from an automatic word aligner). Each issue yields one multi-line note \texttt{build\_oracle\_note(issue)} with bullet lines, in order when the fields are present:

\begin{lstlisting}
- Problematic trace snippet: {trace_quote}
- Relevant source quote: {source_quote}
- Original output quote: {output_quote}
- Why it is problematic: {rationale}
- Use the source sentence to avoid carrying this error into the final translation.
\end{lstlisting}

\begin{algorithmic}[1]
\State \textbf{Oracle-1:} for each targeted issue, call \texttt{from\_trace} with \texttt{record.trace} and a single extra note:
\State \quad \texttt{Oracle correction for one identified issue:} + newline + \texttt{build\_oracle\_note(issue)}
\State \textbf{Oracle-K:} once per record (shared across all targeted issues on that record), call \texttt{from\_trace} with \texttt{record.trace} and extra notes:
\State \quad \texttt{Oracle corrections for all identified issues:}
\State \quad followed by \texttt{build\_oracle\_note($\cdot$)} for every targeted issue on that record, in order
\end{algorithmic}

Both use the shared replay user template; generation is the same as for baseline replay (Section~\ref{app:replay-shared}).

\end{document}

%% file: comments.tex
\newcommand{\ignore}[1]{}

%% file: sections/intro-2.tex
\section{Introduction}

Despite reasoning being applied successfully (à la ``let's think step-by-step'', or through ``thinking-native'' large reasoning models (LRMs)) across canonical NLP tasks, reasoning for machine translation has been met with mixed success. The overwhelming consensus in MT is that reasoning only helps when it is explicitly geared toward translation. For example, a prompt design could involve surfacing or researching translation-relevant knowledge before generation \citep{briakou-etal-2024-translating}, decomposing translation into multiple steps like translating phrases before combining them \citep{he-etal-2024-exploring} or as drafting, refinement, and proofreading \citep{briakou-etal-2024-translating, li2025tactictranslationagentscognitivetheoretic, feng-etal-2025-tear}, or broadly coordinating agentic workflows assigning a role to an agent to conduct steps inducive to translation  \citep{li2025tactictranslationagentscognitivetheoretic, wang-etal-2025-drt}. On the other hand, evaluations of LRMs suggest that off-the-shelf ``native reasoning'' underperforms in MT when compared to direct translation \citep{li-etal-2026-test, rajaee2026unlockingreasoningcapabilitymachine}. Across this body of work, reasoning has been treated instrumentally, basically as a means to obtain better translations. An audit on potential errors in the reasoning is underexplored, and we focus on \textbf{potential errors} present in the reasoning -- whether a reasoning step is correct; if not, whether the misconception propagates to the output; and whether targeted correction of those errors can lead to measurably improved translation quality.

% \textbf{We focus our study on whether reasoning errors could be reliably detected, and test lightweight interventions to probe whether can lead to improved translation quality.
% We also make the argument that reasoning traces could be useful for monolingual users in \textbf{translation verification} -- a signal in helping users determine if a translation is valid, semantically equivalent to the source, and clearly actionable\cb{this is an attempt to make the point that a translation is used for a purpose -- to make decisions, to share, etc.}. % to be used.

In this study, we look to reasoning as our primary object-of-study to understand why they might lead to subpar performance. We find examples of reasoning error that may be propagated to the output (c/f \autoref{fig:motivating-example}). Such errors can arise even in high-resource settings such as English $\rightarrow$ Spanish and English $\rightarrow$ Chinese.
% We also find instances of reasoning-inert but largely useless content in traces that do not enable the models to do translation. 
To automatically detect these errors, we propose and empirically benchmark an LLM-as-a-judge detector, validating them with bilingual human judgments for severity and detectability. Then, we conduct an evaluation to test whether simple, targeted interventions on the detected reasoning errors lead to improved outcomes in the translation. %Finally, we present the detected issues paired with their reasoning traces to monolingual users to assess whether the detected issues can help users with translation verification, and with deciding how to act on the translations.

We guide our study with two central research questions:

\begin{itemize}
\item How frequent are reasoning errors in MT reasoning traces? (\textbf{RQ1})
\item Are detected issues easily mitigated? (\textbf{RQ2}) 
% \item To what extent are these traces detectable and actionable for human auditors? (\textbf{RQ3})
\end{itemize}

\input{figures/figure1}

%% file: figures/figure1.tex
\begin{figure}[t]
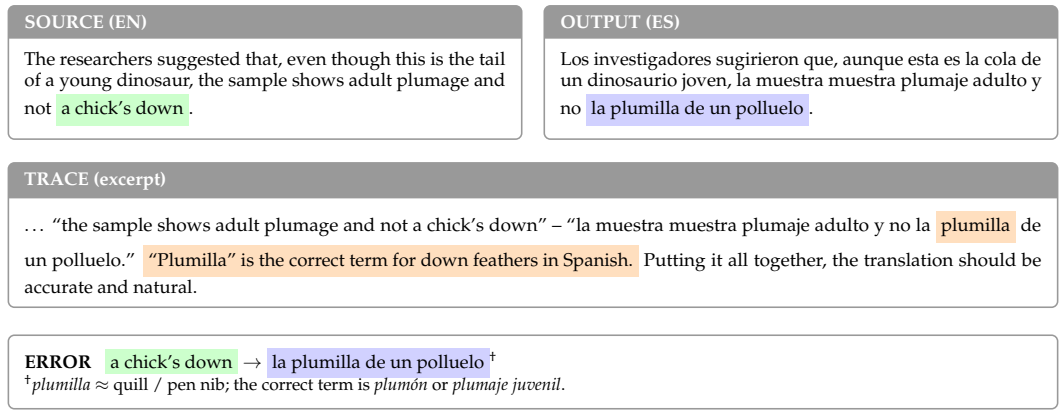

\centering
\scriptsize
\setlength{\fboxsep}{2pt}

\begin{minipage}[t]{0.49\linewidth}
\begin{tcolorbox}[auditbox,title=\textbf{SOURCE (EN)},
  top=2pt,bottom=2pt,left=3pt,right=3pt]
The researchers suggested that, even though this is 
the tail of a young dinosaur, the sample shows adult 
plumage and not \srcHL{a chick's down}.
\end{tcolorbox}
\end{minipage}
\hfill
\begin{minipage}[t]{0.49\linewidth}
\begin{tcolorbox}[auditbox,title=\textbf{OUTPUT (ES)},
  top=2pt,bottom=2pt,left=3pt,right=3pt]
Los investigadores sugirieron que, aunque esta es la 
cola de un dinosaurio joven, la muestra muestra 
plumaje adulto y no \outHL{la plumilla de un polluelo}.
\end{tcolorbox}
\end{minipage}

\vspace{2pt}

\begin{tcolorbox}[auditbox,title=\textbf{TRACE (excerpt)},
  top=2pt,bottom=2pt,left=3pt,right=3pt]
\ldots\ ``the sample shows adult plumage and not a 
chick's down'' -- ``la muestra muestra plumaje adulto 
y no la \trHL{plumilla} de un polluelo.'' 
\trHL{``Plumilla'' is the correct term for down 
feathers in Spanish.} Putting it all together, the 
translation should be accurate and natural.
\end{tcolorbox}

\vspace{2pt}

\begin{tcolorbox}[auditbox,top=2pt,bottom=2pt,
  left=3pt,right=3pt]
\textbf{ERROR}~~
\srcspan{a chick's down} $\rightarrow$ 
\tgtspan{la plumilla de un polluelo}\textsuperscript{\dag}\\[-0.3em]
{\fontsize{6.5}{8}\selectfont 
\textsuperscript{\dag}\emph{plumilla} $\approx$ 
quill / pen nib; the correct term is \emph{plumón} 
or \emph{plumaje juvenil}.}
\end{tcolorbox}

\caption{A reasoning trace decision in Qwen3-8B 
(FLORES, en$\to$es) maps \emph{a chick's down} to 
\emph{plumilla} (quill/pen nib) instead of 
\emph{plumón} (down feathers). The error propagates 
to the output.}
\label{fig:motivating-example}
\end{figure}

%% file: sections/background-2.tex
\section{Background}

\paragraph{Reasoning for machine translation.}
Recent work explores how reasoning can be applied successfully in machine translation, finding that structured and human-like translation strategies that can be decomposed into multiple steps tends to work best. \citet{he-etal-2024-exploring} induces the model to consider translation-related knowledge before producing the final translation. Similarly, \citet{briakou-etal-2024-translating} proposes a pipeline consisting of pre-drafting, drafting, refinement, and proofreading stages. \citet{feng-etal-2025-tear, li2025tactictranslationagentscognitivetheoretic} introduce frameworks that consist of collaborative agents that collaborate in order to generate the best final translation, primarily consisting of distinct candidate generation + verification roles. In contrast with what reasoning-like behavior can induce better translation output, \citet{li-etal-2026-test, rajaee2026unlockingreasoningcapabilitymachine} both find that ``native reasoning'' underperforms when compared to the direct translation approach.

\paragraph{Faithful reasoning and trace sensitivity.}
Our faithfulness tests are most closely related to work on reasoning faithfulness, which asks whether a model's output is causally attributable to its intermediate reasoning \citep{lanham2023measuringfaithfulnesschainofthoughtreasoning}. 
With multiple-choice Question-Answering benchmarks,
\cite{lanham2023measuringfaithfulnesschainofthoughtreasoning} directly intervenes on Chain-of-Thought reasoning by perturbing traces and measures model faithfulness by the extent to which the model’s accuracy changes. Interestingly, they find that larger models are 
often less faithful: the final answer survives even substantial trace perturbations, suggesting the model is not always faithful to the reasoning in all settings. 

\citet{paul-etal-2024-making} sharpens this perspective this intuition 
via causal mediation analysis, 
modeling the trace $R$ as a mediator between input $X$ and output $Y$. They decompose the total effect into a \textbf{direct effect} ($X \to Y$, bypassing $R$) and an 
\textbf{indirect effect} ($X \to R \to Y$), measured by intervening on both the input 
(swapping in a counterfactual question $X_1$) and the trace (swapping in reasoning $R_1$ generated for $X_1$). 
Across several reasoning-oriented benchmarks (commonsense \& math) they find that indirect effects decrease as reasoning traces get larger. They also show 
that none of the tested models exhibit consistently high indirect effects, indicating that LLMs do not reliably 
use their stated reasoning when producing answers. More recently, \citet{chen2025reasoningmodelsdontsay} study  faithfulness in the extended thinking traces of LRMs, and find that models frequently omit mention of hints they relied on, with this type of faithfulness decreasing on harder tasks. \cite{zhou2024languagemodelsperformrobust} further shows that models are vulnerable to inaccurate noisy rationales in-context, with noisy reasoning degrading downstream accuracy. Together, these observations motivate our setting in MT: if models inconsistently condition on their own reasoning in more verifiable tasks, the questions of whether this extends to the richer output space of NLG tasks like translation, where the output may or may not reflect many decisions claimed in the reasoning, enabling a richer analysis that is not possible in the settings with verifiable answers.

%% file: sections/framework.tex
\section{Methodology for Quantifying Reasoning Errors}

\subsection{Defining reasoning errors}
\input{tables/matrix}

Prior work on chain-of-thought faithfulness asks whether a model's stated reasoning influences the final output \citep{lanham2023measuringfaithfulnesschainofthoughtreasoning, 10.5555/3666122.3669397, vonrecum2026reasoningllmsrobustinterventions}. This work typically intervenes on reasoning chains, by truncating, paraphrasing, injecting errors, and measuring robustness of model responses to these perturbations by assessing whether the model's answer changes in the intended way. When it does not, the model is deemed unfaithful to the input reasoning, as canonically defined \citep{jacovi-goldberg-2020-towards}. 
% There is also work in interpretability that assesses whether reasoning is causally faithful in the parametric sense \citep{tutek-etal-2025-measuring}.

% We adapt a different but related framing for the task of machine translation by first asking whether there are errors in the trace, and then by intervening on these errors and assessing whether these interventions correct any errors, if exists, in the output.

We propose 3 consistency categories connected to distinct threads of prior work in evaluation for NLG, reasoning verification, and CoT faithfulness, which are referred in \autoref{tab:error_taxonomy}.

\input{tables/motivation_for_categories}

\subsection{Detecting reasoning errors}
\label{sec:annotation_protocol}

Inspired by the evaluation axes of the general reasoning trace evaluation taxonomy proposed by \citet{lee-hockenmaier-2025-evaluating}, we operationalize this for machine translation. \textsc{Source--Reasoning} and 
\textsc{Reasoning-Internal} instantiate the 
\emph{factuality} and \emph{validity/coherence} categories.  The closest analogy to \textsc{Reasoning--Output} relates to both the 
\emph{utility} axis of \citet{lee-hockenmaier-2025-evaluating} 
and to chain-of-thought faithfulness 
\citep{lanham2023measuringfaithfulnesschainofthoughtreasoning}.

We define a structured annotation protocol that decomposes each trace into individual decision steps simply via sentence tokenization. It then checks the consistency of each reasoning step with the preceding reasoning trace, the source input and the model output. In terms of granularity of annotation, we choose to annotate at the reasoning step-level for both practical reasons and inspired by work in Chain-of-Thought faithfulness, which frequently operates at the step-level to support targeted interventions, as we would like to do \citep{prasad-etal-2023-receval, jacovi-etal-2024-chain}. 

\paragraph{Accounting for self-correction.} Reasoning traces are generative processes: models often explore, reconsider, and self-correct within a single trace. \citet{lanham2023measuringfaithfulnesschainofthoughtreasoning} finds that models can recover from errors introduced earlier in their reasoning chains, but recovery is not guaranteed and depends on error type and task. To account for this, we distinguish between two severity levels. An issue is marked \textbf{\textsc{Error}} if the incorrect decision persists through the trace. An issue is marked \textbf{\textsc{Fixed-later}} if the trace initially makes an error but then corrects it at a later point in the trace. This distinction is important because it separates errors that persists from natural reasoning errors that may arise from exploration, and discarded by the end. We discard \textbf{\textsc{Fixed-later}} from our analysis going forward. % We hypothesize that both matter for a user, but for different reasons: \textsc{Errors} are relevant to output quality, while \textsc{Fixed-later}s are relevant to the trace's value as a verification aid, since a user reading the trace would see the incorrect reasoning.

\paragraph{Error Detection Protocol/Implementation}

We implement the protocol with an LLM-as-a-judge (GPT-5.2). The judge receives three inputs: the source sentence, the sentence-tokenized reasoning trace, and the final translation. For each issue it identifies, it returns a structured annotation with: (i) the category (\textsc{Input--Reasoning}, \textsc{Reasoning--Output}, or \textsc{Reasoning--Internal}); (ii) the index into the trace sentence; and (iii) and a short rationale.

% recent work on scoring step-by-step reasoning?

% \subsection{Case study: English$\rightarrow$Spanish}
\label{sec:case_study}

We sample $k = 5$ judgments per trace with temperature set to 0.4. For each independently sampled run, we then aggregate by category and reasoning step index  and keep only the issues with at least 3 judgments referring to it (majority vote, $\geq$ 3/5). We test on 7 language directions: en $\rightarrow$ \{es,ja,fr,de,yue,ur,zh\}, sourced from the FLORES parallel test set \citep{goyal-etal-2022-flores}.  
% \cb{TODO: add appendix detailing the prompts used for llm-as-a-judge and statistics on intra-annotator agreement given k samples}

% \cb{TODO: add section for measuring surprisal of detected errors, if we see evidence of it}

\paragraph{Detected reasoning errors}

\begin{table*}[t]
\centering
\footnotesize
\setlength{\tabcolsep}{3pt}
\textbf{Annotation summary by reasoning model}
\vspace{0.25em}
\begin{tabular*}{\textwidth}{@{\extracolsep{\fill}}llrrrrr@{}}
\toprule
Model & Language pair & \# Samples & \# w/ errors & Error rate & Avg.\ steps / trace & Avg.\ errors / sample \\
\midrule
\multicolumn{7}{l}{\textit{Qwen3-8B}} \\
\midrule
 & en$\rightarrow$es & 100 & 56 & 56.0\% & 39.12 & 1.40 \\
 & en$\rightarrow$zh & 100 & 54 & 54.0\% & 32.69 & 1.86 \\
 & en$\rightarrow$ja & 100 & 86 & 86.0\% & 40.88 & 3.51 \\
 & en$\rightarrow$fr & 100 & 73 & 73.0\% & 42.31 & 2.81 \\
 & en$\rightarrow$de & 100 & 81 & 81.0\% & 45.64 & 2.94 \\
& en$\rightarrow$ur & 100 & 97 & 97.0\% & 38.26 & 6.40 \\
& en$\rightarrow$yue & 100 & 63 & 63.0\% & 36.25 & 2.30 \\

\midrule
 & all & 700 & 510 & 72.8\% & 39.31 & 3.03 \\
\midrule
\multicolumn{7}{l}{\textit{Ministral-3-8B}} \\
\midrule
 & en$\rightarrow$es & 100 & 60 & 60.0\% & 21.59 & 2.24 \\
 & en$\rightarrow$zh & 100 & 65 & 65.0\% & 17.63 & 3.18 \\
 & en$\rightarrow$ja & 100 & 93 & 93.0\% & 22.20 & 3.93 \\
 & en$\rightarrow$fr & 100 & 74 & 74.0\% & 16.84 & 2.72 \\
 & en$\rightarrow$de & 100 & 69 & 69.0\% & 20.34 & 3.39 \\
 & en$\rightarrow$ur & 100 & 100 & 100.0\% & 16.40 & 5.68 \\
& en$\rightarrow$yue & 100 & 100 & 100.0\% & 17.12 & 3.74 \\
\midrule
 & all & 700 & 561 & 80.1\% & 19.72 & 3.55 \\
\bottomrule
\end{tabular*}
\caption{LLM-as-a-judge reasoning errors on FLORES devtest, $n{=}100$ per direction.} 
\label{tab:annotation_summary}
\end{table*}

\autoref{tab:annotation_summary} shows a summary of all detected errors in the reasoning traces generated by Qwen3-8B and Ministral-3-8B. We see a large proportion of samples (72.8\% on average for Qwen3-8B, and 80.1\% on average for Ministral-3-8B) across languages with at least one reasoning error, with on average 3.03 and 3.55 average errors per sample. These rates are high enough to raise the question of whether the LLM-as-a-judge is identifying genuine errors, or is pedantically flagging reasoning steps. To distinguish this, we validate a subset of 
the detected issues against bilingual human annotators (\S\ref{subsec:human_eval}) in two language pairs: en$\rightarrow$es (high-resource with the smallest avg. errors / sample), and  en$\rightarrow$ur (low-resource with the highest avg. errors / sample). 

% \paragraph{Bilingual human validation}
% freitag-etal-2021-experts
% graham-etal-2013-continuous

\subsection{Human validation}\label{subsec:human_eval}

We conduct a validation via bilingual human annotators to compute the precision of the automatic LLM-as-a-judge results on error detection, using Qwen3-8B translations for English$\rightarrow$Spanish, and English$\rightarrow$Urdu. For each language pair, we sample 30 FLORES devtest examples. Each example is a triplet $(x,t,y)$, where $x$ is the English source sentence, $t$ is the model reasoning trace, and $y$ is the final translation. Each sampled example  from FLORES yields one item-level annotation for translation quality. In addition, each sample yields one or more reasoning errors in $t$, which are all annotated for error validity. All items are annotated independently by three bilingual annotators per language pair. We recruit bilingual language specialists rather than general crowd workers, following conclusions that judgments involving error analysis by experts is more reliable than judgments from non-specialists \citep{freitag-etal-2021-experts}.

\paragraph{Phase 1: item-level output validity.}
In the first phase, annotators see only the source sentence $x$ and the final translation $y$. They answer the following question:

\begin{quote}
\textbf{Does the translation preserve the important meaning of the source sentence?}
\end{quote}

Annotators choose one of three labels:
\begin{itemize}
    \item \textsc{OK}: no critical meaning error is present; minor fluency or stylistic differences are acceptable;
    \item \textsc{Not OK}: the translation contains at least one meaning error that could mislead a reader;
    \item \textsc{Unsure}: I cannot confidently judge the item.
\end{itemize}

For items labeled \textsc{Not OK}, annotators additionally indicate the shortest source span whose meaning is not preserved and the shortest target-language span containing the error. These labels serve two purposes in the analysis. First, they provide an estimate of final-output correctness that is independent of the reasoning trace. Second, when combined with issue-level trace validation below, they allow us to distinguish \emph{right answer, wrong reason} cases (valid output despite a validated trace error) from \emph{wrong answer, wrong reason} cases (validated trace error that contributes to an invalid output).

\paragraph{Phase 2: issue-level validation of detector-flagged trace spans.}
In the second phase, annotators validate individual issues proposed by the detector. For each flagged issue, annotators see the source sentence $x$, the final translation $y$, and the sentence-tokenized reasoning trace $t$, with the detected reasoning error highlighted. 

Specifically, annotators answer the following questions:

\begin{enumerate}
    \item \textbf{Does the highlighted reasoning span contain an error?} \\
    Response options: 
     \\ - \textsc{Yes} -- the highlighted reasoning contains a clear mistake (wrong meaning, wrong claim, a contradiction, etc.), \\
    - \textsc{No} -- The highlighted reasoning is correct or a reasonable interpretation, \\
    - \textsc{Borderline} -- the reasoning is imprecise or debatable but not clearly wrong. \\

    \item \textbf{How confident are you in your judgment above?}  \\
    Response options: \\ - \textsc{CONFIDENT} — I am sure of my answer. \\ - \textsc{SOMEWHAT CONFIDENT} — I think my answer is right but it requires specialized knowledge I'm not fully certain about. \\ - \textsc{NOT CONFIDENT} — This requires detailed language knowledge I'm not fully certain about. \\

    \item \textbf{ Is this reasoning step reflected in the final translation? } \\
    Response options: \\
    - \textsc{Yes} -- I see the decision or claim reflected in the final translation,\\ - \textsc{No} -- The final translation seems to not have the decision or claim reflected in it, \\ - \textsc{Not applicable} -- I cannot determine the connection. \\

    \item \textbf{Which of the following categories apply to this highlighted reasoning step? (multi-select)} \\
    % Source misinterpretation - 
    Response options: \\ - \textsc{Source misinterpretation} -- The reasoning step is making a wrong claim or decision about the source. The reasoning misunderstands or misrepresents something in the English source (e.g., wrong word meaning, hallucinated information not in the source), \\ - \textsc{Internal Contradiction} -- The reasoning step contradicts its context. contradicts itself or contains circular/incoherent logic, \\ - \textsc{No issue} -- This reasoning step makes sense to me in this context. \\ - \textsc{Other / unsure}. 
\end{enumerate}

% \paragraph{Reported metrics.}
% For each question, we aggregate the three annotator responses by majority vote. If no response obtains a majority, or if the majority label is an uncertainty label (e.g., \textsc{Unclear} or \textsc{Not applicable}), we mark that question for that instance as unresolved and report the rate of unresolved cases separately. We then report four classes of statistics.

First, we report \textbf{overall detector precision}, defined as the proportion of flagged issues is validated as a reasoning error by the corresponding majority human judgment.  Second, we compare reasoning-level validation with Phase~1 item-level \textsc{OK}/\textsc{Not OK} judgments. % This comparison lets us quantify how often the trace has a genuine problem even when the final output is acceptable (\emph{right answer, wrong reason}), versus how often a validated trace error is also associated with a meaning error in the final translation (\emph{wrong answer, wrong reason}). 

% \paragraph{Results.}
% \cb{TODO add table for reporting agreement across languages tested. }
\input{tables/human_eval_results}
% \subsection{Bilingual human validation.}
% \input{figures/human_eval}

% English--Urdu human validation discussion
% my own notes: - sample_aggregation.csv: all English--Urdu rows
%
% - sample_aggregation.csv rows 2--31; ties at rows 12 and 16
% - annotator_summary.csv rows 2--4
% - category_summary.csv rows 2--3
% - raw_sample_annotations.csv rows for case studies:
%     * Qwen3-8B_en-ur_001: rows 2, 32, 58
%     * Qwen3-8B_en-ur_002: rows 3, 33, 59
%     * Qwen3-8B_en-ur_026: rows 27, 57, 83
% - issue_aggregation.csv:
%     * Qwen3-8B_en-ur_001: rows 3--5
%     * Qwen3-8B_en-ur_014: rows 75--86

\paragraph{English--Urdu.}
Across a sample of 30 English$\rightarrow$Urdu (86 sample-level annotations and 541 issue-level annotations), at the sample level, 28/30 items receive a majority translation fidelity error with a translation \textsc{NOT OK} judgment, with mean confidence high (0.819--1.000). 

In terms of what the detector flagged as reasoning errors, they are validated by humans. Precision is 93.1\% averaged across 3 annotators, with high confidence (94.0\%). 41.1\% of detected reasoning errors are judged to be reflected in the final output, indicating that a substantial fraction of the detected trace errors are not merely local reasoning noise but also impact the downstream output. Source misinterpretations are also more likely to propagate to the final Urdu translation: among error annotations, the reflection-\textsc{Yes} rate is 44.9\% for source misinterpretation, compared to 20.2\% for internal contradiction. This suggests that the detector is useful when it flags these errors, and that these errors propagate to poor English $\rightarrow$ Urdu outputs. Taken together, the English$\rightarrow$Urdu results paint a consistent picture: final translations are deemed not semantically faithful to the source, and bilingual annotators agree with the resulting errors, indicating high precision.
% strong precision 

\paragraph{English--Spanish.}
The English$\to$Spanish results 
(\autoref{tab:human_validation_summary}) contrast sharply. 80.0\% of samples receive a majority \textsc{Yes} on translation correctness: the Spanish translations are largely faithful to the English source. The judge flags 58 reasoning errors in these samples, but annotators confirm only 50.0\% of them, with 37.9\% rejected as 
non-errors. Crucially, annotators are more confident when rejecting the judge's flags (97.3\% confidence on \textsc{No}) than when confirming them (88.9\% on \textsc{Yes}). 

Among the free-text responses, annotators note cases where the judge objects to a reasonable translation choice, flags an ambiguity the source itself does 
not resolve (e.g., whether ``garage'' means a workshop or a literal garage), or claims a grammatical rule stricter than what a native speaker would expect. Among the errors annotators do confirm, the output propagation rate is high: 84.0\% are  decided to have been reflected in the final output, compared to 41.1\% in Urdu. This suggests  that the judge's threshold in Spanish is high enough that only genuinely consequential errors are flagged as errors, and those tend to matter for the translation.

% \cb{[TODO: Add the parallel English$\rightarrow$Spanish discussion here. summarize (i) sample-level translation correctness, (ii) reasoning-error precision and annotator confidence, (iii) the category breakdown, and (iv) one or two illustrative case studies from the raw and issue-level annotations.]}
% \cb{There should also be a table.}

% \cb{ Placeholder for human validation plan using translators from Upwork. \footnote{https://docs.google.com/document/d/15fvjQ2O-pJo9WXAxqyOluYPX6EJrEKD3_ENRR_Xn-I8/edit?tab=t.0#heading=h.uqqw3n1r3ntq}}

%% file: tables/matrix.tex
% Preamble:
% \usepackage{booktabs,tabularx,array,amssymb}

\begin{table}[t]
\centering
\small
\renewcommand{\arraystretch}{1.2}
\begin{tabularx}{\textwidth}{
    >{\raggedright\arraybackslash}p{3.2cm}
    >{\raggedright\arraybackslash}X
    >{\raggedright\arraybackslash}X
}
\toprule
& \textbf{Output Valid (O\cmark )} & \textbf{Output Invalid (O\xmark )} \\
\midrule

% \newcommand{\cmark}{\textcolor{green!60!black}{\ding{51}}}  % ✓
% \newcommand{\xmark}{\textcolor{red}{\ding{55}}}             % ✗
% \rowcolor{green!8}

\textbf{Reasoning Valid (R\cmark)}
& \textbf{Potentially aligned}: trace has no detected issues and the translation has no detected issues.
& \textbf{Silent translation error}: translation is wrong, but the trace does not contain detectable issues. \\

\addlinespace

\rowcolor{red!8}
\textbf{Reasoning Invalid (R\xmark)}
& \textbf{Right answer, wrong reason}: translation is correct, but the trace contains a detected issue. This may cause reduced trust in good translations.
& \textbf{Wrong answer, wrong reason}: translation is incorrect, and trace has semantic divergences. Highlighting this may help users reject imperfect translations. \\

% \textbf{Reasoning Invalid (R\xmark)}
% & \textbf{Right answer, wrong reason}: translation is correct, but the trace contains a detected issue. This may cause reduced trust in good translations.
% & \textbf{Wrong answer, wrong reason}: translation is incorrect, and trace has semantic divergences. Highlighting this may help users reject imperfect translations. \\

\bottomrule
\end{tabularx}
\caption{A 2$\times$2 matrix framing the validity of reasoning and translation outputs. In this work, we analyze the detection of invalid outputs, conditioned on invalid reasoning R\xmark.}
% on the detection and mitigation of the two categories where reasoning is invalid R\xmark.} %in the bottom, where reasoning is invalid.}
\label{tab:trace_output_2x2}
\end{table}

%% file: tables/motivation_for_categories.tex
\begin{table*}[t]
\centering
\small
\renewcommand{\arraystretch}{1.4}
\begin{tabularx}{\textwidth}{@{} l p{2.6cm} p{4.6cm} p{4.6cm} @{}}
\toprule
\textbf{Category} & \textbf{Question} & \textbf{Example} & \textbf{Connection to \cite{lee-hockenmaier-2025-evaluating}} \\
\midrule
\begin{tabular}[t]{@{}l@{}}
\textsc{Input--}\\
\textsc{Reasoning}
\end{tabular}
&
Does the trace make claims supported by the source sentence?
&
\textbf{Source}: ``The \errtxt{bug} crashed the whole system.''
\textbf{Step~3}: ``The word bug here means \errtxt{a biological organism}; 
I'll translate it as \errtxt{\emph{bacteria}}.''
\textbf{Explanation:} The trace selects the wrong sense of \emph{bug}; the source context (software) does not support this interpretation.
&
In the taxonomy of \citet{lee-hockenmaier-2025-evaluating}, this corresponds most directly to \textbf{factuality}, especially groundedness with respect to the source. Analogous to \textbf{groundedness} in NLG evaluation \citep{honovich-etal-2022-true,maynez-etal-2020-faithfulness}: just as a summary can hallucinate content absent from a source document, a trace can hallucinate translation equivalences.
\\[6pt]
\begin{tabular}[t]{@{}l@{}}
\textsc{Reasoning--}\\
\textsc{Output}
\end{tabular}
&
Does the output reflect the decisions stated in the trace?
&
\textbf{Step~8}: ``The phrase `pretending to listen' maps 
naturally to \errtxt{\emph{haciendo como si estuviera 
escuch\'{a}ndonos}}.''
\textbf{Step~9}: ``I'm confident in this phrasing. Finalizing the translation.''
\textbf{Output}: ``Ella estaba \errtxt{\emph{haci\'{e}ndose pasar 
por si estuviera escuch\'{a}ndonos}}.''
\textbf{Explanation:} The output uses a different phrase than the one committed to in Steps~8--9.
&
In the taxonomy of \citet{lee-hockenmaier-2025-evaluating}, this aligns most closely with \textbf{utility}: whether a reasoning step actually contributes to the final answer. Prior work on CoT faithfulness asks a related question: whether reasoning reflects the model's actual process \citep{lanham2023measuringfaithfulnesschainofthoughtreasoning,jacovi-goldberg-2020-towards}. %This category instead measures the apparent \textbf{causal influence} of stated decisions on the output, closer to mediation analysis \citep{paul-etal-2024-making}.
\\[6pt]
\begin{tabular}[t]{@{}l@{}}
\textsc{Reasoning-}\\
\textsc{Internal}
\end{tabular}
&
Is the trace internally consistent and coherent?
&
\textbf{Step~5}: ``\emph{bank} means \errtxt{\emph{el banco}}.'' 
\textbf{Step~8}: ``Wait---since it is next to `river', it should be \errtxt{\emph{la orilla}}.'' 
\textbf{Step~9}: ``Alright, going with \errtxt{\emph{el banco}} as decided earlier.''
\textbf{Explanation:} Step~9 reverts to Step~5 without acknowledging the correction in Step~8.
&
In the taxonomy of \citet{lee-hockenmaier-2025-evaluating}, this aligns with both \textbf{coherence} and \textbf{validity}. Coherence is concerned with whether a reasoning step's preconditions are established consistently, while validity is concerned with whether the step contains a logical inconsistency. Draws also from detection of self-contradictory reasoning in LLM outputs \citep{liu-etal-2024-self-contradictory}.
\\
\bottomrule
\end{tabularx}
\caption{Error categories for reasoning errors in machine translation, grounded in the reasoning-trace evaluation taxonomy of \citet{lee-hockenmaier-2025-evaluating}. \errtxt{Highlighted spans} in the example indicate the error span to focus on.}
\label{tab:error_taxonomy}
\end{table*}

%% file: tables/human_eval_results.tex
\begin{table*}[t]
\centering
\small
\setlength{\tabcolsep}{6pt}
\renewcommand{\arraystretch}{1.15}
\begin{tabular}{@{}lcc@{}}
\toprule
& English$\to$Urdu & English$\to$Spanish \\
\midrule
\multicolumn{3}{@{}l}{\textit{Study size}} \\
% Annotators & 3 & 3 \\
Samples annotated & 30 & 30 \\
Issues annotated & 189 & 58 \\
\addlinespace
\midrule

\multicolumn{3}{@{}l}{\textit{Translation correctness (majority vote per sample)}} \\
\textsc{Yes} (no major error) & 0/30 (0.0\%) & 24/30 (80.0\%) \\
\textsc{No} (major error) & 28/30 (93.3\%) & 5/30 (16.7\%) \\
\textsc{Tie} & 2/30 (6.7\%) & 1/30 (3.3\%) \\
\addlinespace
\midrule

\multicolumn{3}{@{}l}{\textit{Reasoning-error validation (majority vote per issue)}} \\
\textsc{Yes} only & 176/189 (93.1\%) & 29/58 (50.0\%) \\
\textsc{Yes} + \textsc{Borderline} & 179/189 (94.7\%) & 31/58 (53.4\%) \\
\textsc{No} & 3/189 (1.6\%) & 22/58 (37.9\%) \\
\textsc{Tie} & 7/189 (3.7\%) & 5/58 (8.6\%) \\
\addlinespace
\midrule

\multicolumn{3}{@{}l}{\textit{Error reflected in output? (among validated errors only)}} \\
\textsc{Yes} & 41.1\% & 84.0\% \\
\textsc{No} & 56.6\% & 16.1\% \\
\textsc{Unsure} & 2.3\% & 0.0\% \\
\addlinespace
\midrule

\multicolumn{3}{@{}l}{\textit{Annotator confidence}} \\
Mean confidence & 90.5\% & 91.4\% \\
On \textsc{Yes} judgments & 94.0\% & 88.9\% \\
On \textsc{No} judgments & 72.8\% & 97.3\% \\
\addlinespace
\bottomrule
\end{tabular}
\caption{Human validation summary for 
English$\to$Urdu and English$\to$Spanish. 
Translation correctness and reasoning-error 
validation are reported at the sample and issue 
level respectively, using majority vote across 
three annotators. Error reflection rates are 
computed over annotations labeled \textsc{Yes} 
or \textsc{Borderline}.}
\label{tab:human_validation_summary}
\end{table*}

%% file: sections/intervention.tex
\section{Interventions on Reasoning}

We test whether detected reasoning errors causally influence the model's translation. The core idea follows the 
interventionist logic of causal mediation analysis \citep{paul-etal-2024-making}: if 
modifying a trace step changes the output, that step has causal influence on the translation. Starting from the original triplet $(x,t,y)$ where $t$ and $y$ are generated in a model's thinking mode, we ``re-play'' the trace on the same model with thinking disabled, while modifying $t \mapsto t'$ and injecting $t'$ as context to obtain a translation $y'$. %The caveat is that this is not akin to generating the thinking trace as is, ... 

% \begin{quote}
% \small
% \texttt{[System]} \emph{You are a machine 
% translation assistant.}\\
% \texttt{[User]} \emph{$\langle$task instruction$\rangle$ 
% Source: $x$ \textbackslash n Reasoning trace: $t'$ 
% \textbackslash n Use the reasoning trace when deciding on the translation. Return only the final translation.}
% \end{quote}

% In the original generation, $t$ is produced autoregressively within a 
% \texttt{<think>...</think>} block before $y$ is generated. In this setting, 
% $t'$ appears in the user turn as provided context in the prompt. This 
% means the token sequence differs from the original: the trace occupies different absolute positions and is no longer surrounded by special `think` tokens.
% % Consequently, the replay does not recover the original generation dynamics. Instead what we are measuring is how the model 
% translates when conditioned a given reasoning 
% trace as input context. This distinction  
% parallels the difference between 
% observational CoT analysis (measuring what 
% happens during generation) and 
% interventional analysis (measuring what changes when the trace is modified), the latter being the 
% standard for causal claims about CoT influence 
% \citep{lanham2023measuringfaithfulnesschainofthoughtreasoning}.

\subsection{Types of Interventions}
We study six interventions with increasing expected strength. We construct controlled transformations $t \mapsto t'$ with just a modification on the detected reasoning error for the first three, and we consider the last three ``oracle''-level interventions, since they construct $t'$ guided by the reference, instead of the detected issue. 
\begin{itemize}
  \item \textbf{Hedging:} prepend hedging language to the reasoning error  (e.g., \emph{``I'm not sure, but\ldots''}).

  \item \textbf{Removal:} remove the detected reasoning error from the trace.
  \item \textbf{Re-reasoning after removal:} remove the error span and regenerate the remainder of the trace from that point onward, allowing the model to re-derive downstream reasoning.

\item \textbf{Hindsight:} use the model to generate a reasoning trace to translate from the source to the reference, then re-play with that reasoning trace.

  \item \textbf{Oracle-$K$:} $K$ phrase-level hints covering the full source--reference alignment.
  \item \textbf{Oracle-1:} a single hint mapping the error-relevant source phrase to its reference translation.
\end{itemize}

% \paragraph{Oracle upper bounds.}
% In these baselines, we inject  hints based on the reference to serve as upper bound baselines.
% \begin{itemize}

% \end{itemize}

% The oracle interventions represent upper bounds. They are not appropriate reasoning strategies that might mitigate potential issues in ``natural'' reasoning traces for the model, but they highlight how the model would fare given ``gold'' hints during reasoning.
% had the model is guided by references in the reasoning.

% strong reasoning information for task, guided by translation references.

\input{tables/replay_deux2}

%% file: tables/replay_deux2.tex
% % % % % % Combined intervention table across all language pairs
% % % % % \begin{table*}[t]
% % % % % \centering

\begin{table*}[t]
\centering
\footnotesize
\setlength{\tabcolsep}{4pt}
\renewcommand{\arraystretch}{1.05}
\begin{tabular}{@{}lccc ccc@{}}
\toprule
& \multicolumn{3}{c}{Qwen3-8B} & \multicolumn{3}{c}{Ministral-3-8B} \\
\cmidrule(lr){2-4}\cmidrule(lr){5-7}
Intervention & Resolved / total & Rate $\uparrow$ & $\Delta$ COMET $\uparrow$ & Resolved / total & Rate $\uparrow$ & $\Delta$ COMET $\uparrow$ \\
\midrule
\multicolumn{7}{@{}l}{\textbf{en-de}} \\
hedging & 2 / 149 & 1.3\% & 0.0000 & 48 / 116 & 41.4\% & -0.0010 \\
hindsight & 125 / 149 & \textbf{83.9\%} & \textbf{+0.0088} & 103 / 116 & \textbf{88.8\%} & +0.0001 \\
oracle-1 & 47 / 149 & 31.5\% & +0.0038 & 78 / 116 & 67.2\% & +0.0006 \\
oracle-k & 45 / 149 & 30.2\% & +0.0025 & 77 / 116 & 66.4\% & \textbf{+0.0027} \\
removal & 5 / 149 & 3.4\% & +0.0008 & 59 / 116 & 50.9\% & +0.0006 \\
\shortstack[l]{re-reason} & 75 / 149 & 50.3\% & -0.0024 & 80 / 116 & 68.9\% & -0.0715 \\
\midrule
\multicolumn{7}{@{}l}{\textbf{en-es}} \\
hedging & 0 / 82 & 0.0\% & 0.0000 & 55 / 101 & 54.5\% & +0.0003 \\
hindsight & 67 / 82 & \textbf{81.7\%} & -0.0039 & 88 / 101 & \textbf{87.1\%} & -0.0013 \\
oracle-1 & 17 / 82 & 20.7\% & -0.0009 & 83 / 101 & 82.2\% & +0.0023 \\
oracle-k & 19 / 82 & 23.2\% & -0.0006 & 83 / 101 & 82.2\% & \textbf{+0.0067} \\
removal & 5 / 82 & 6.1\% & -0.0010 & 62 / 101 & 61.4\% & +0.0008 \\
\shortstack[l]{re-reason} & 42 / 82 & 51.2\% & -0.0121 & 71 / 101 & 70.3\% & -0.0777 \\
\midrule
\multicolumn{7}{@{}l}{\textbf{en-fr}} \\
hedging & 9 / 149 & 6.0\% & -0.0012 & 67 / 137 & 48.9\% & +0.0001 \\
hindsight & 114 / 149 & \textbf{76.5\%} & \textbf{+0.0119} & 121 / 137 & \textbf{88.3\%} & \textbf{+0.0055} \\
oracle-1 & 48 / 149 & 32.2\% & +0.0002 & 103 / 137 & 75.2\% & +0.0010 \\
oracle-k & 57 / 149 & 38.3\% & +0.0012 & 102 / 137 & 74.5\% & +0.0030 \\
removal & 10 / 149 & 6.7\% & -0.0013 & 76 / 137 & 55.5\% & -0.0007 \\
\shortstack[l]{re-reason} & 93 / 149 & 62.4\% & -0.0090 & 89 / 137 & 65.0\% & -0.1156 \\
\midrule
\multicolumn{7}{@{}l}{\textbf{en-ja}} \\
hedging & 9 / 200 & 4.5\% & +0.0002 & 94 / 214 & 43.9\% & +0.0064 \\
hindsight & 147 / 200 & \textbf{73.5\%} & \textbf{+0.0110} & 179 / 214 & \textbf{83.6\%} & \textbf{+0.0254} \\
oracle-1 & 56 / 200 & 28.0\% & +0.0020 & 136 / 214 & 63.6\% & -0.0067 \\
oracle-k & 54 / 200 & 27.0\% & +0.0011 & 129 / 214 & 60.3\% & -0.0011 \\
removal & 13 / 200 & 6.5\% & -0.0005 & 101 / 214 & 47.2\% & +0.0026 \\
\shortstack[l]{re-reason} & 101 / 200 & 50.5\% & -0.0003 & 111 / 214 & 51.9\% & -0.1607 \\
\midrule
  \multicolumn{7}{@{}l}{\textbf{en-ur}} \\
  hedging & 2 / 399 & \textbf{0.5\%} & -0.0009 & 42 / 210 & 20.0\% & +0.0030 \\ 
  hindsight & 1 / 399 & 0.3\% & \textbf{+0.1066} & 74 / 210 & 35.2\% & \textbf{+0.0892} \\      
  oracle-1 & 0 / 399 & 0.0\% & +0.0048 & 68 / 210 & 32.4\% & +0.0008 \\               
  oracle-k & 0 / 399 & 0.0\% & +0.0066 & 62 / 210 & 29.5\% & +0.0004 \\                
  removal & 1 / 399 & 0.3\% & -0.0002 & 87 / 210 & 41.4\% & +0.0177 \\               
  \shortstack[l]{re-reason} & 1 / 399 & 0.3\% & -0.0124 & 99 / 210 & \textbf{47.1\%} & -0.1975 \\                          
  \midrule                                   
  \multicolumn{7}{@{}l}{\textbf{en-yue}} \\   hedging & 0 / 135 & 0.0\% & 0.0000 & 12 / 130 & 9.2\% & +0.0004 \\                  hindsight & 0 / 135 & 0.0\% & \textbf{+0.0131} & 17 / 130 & 13.1\% & \textbf{+0.0194} \\       oracle-1 & 3 / 135 & \textbf{2.2\%} & +0.0041 & 9 / 130 & 6.9\% & -0.0039 \\    oracle-k & 1 / 135 & 0.7\% & +0.0004 & 9 / 130 & 6.9\% & -0.0041 \\                 removal & 0 / 135 & 0.0\% & +0.0002 & 7 / 130 & 5.4\% & +0.0019 \\                \shortstack[l]{re-reason} & 0 / 135 & 0.0\% & -0.0156 & 25 / 130 & \textbf{29.2\%} & -0.0938 \\                           
  \midrule  

\multicolumn{7}{@{}l}{\textbf{en-zh}} \\
hedging & 1 / 94 & 1.1\% & -0.0001 & 72 / 121 & 59.5\% & -0.0070 \\
hindsight & 68 / 94 & \textbf{72.3\%} & -0.0012 & 106 / 123 & \textbf{86.2\%} & \textbf{+0.0089} \\
oracle-1 & 25 / 94 & 26.6\% & -0.0012 & 87 / 123 & 70.7\% & -0.0033 \\
oracle-k & 36 / 94 & 38.3\% & +0.0004 & 79 / 123 & 64.2\% & -0.0132 \\
removal & 6 / 94 & 6.4\% & -0.0001 & 73 / 123 & 59.3\% & +0.0026 \\
\shortstack[l]{re-reason} & 47 / 94 & 50.0\% & -0.0102 & 82 / 123 & 66.7\% & -0.0958 \\
\bottomrule
\end{tabular}
\caption{Intervention results grouped by language pair. Bold marks the highest resolution rate and the highest $\Delta$ COMET within each language section. Positive $\Delta$ COMET values are shown with an explicit + sign. Resolved / total preserves the model-specific denominators from the original results.}
\label{tab:intervention_results_sectioned}
\end{table*}

%% file: sections/intervention2.tex
\subsection{Intervention results}\label{subsec:intervention-results}

\autoref{tab:intervention_results_sectioned} reports error resolution rates 
and COMET deltas across five language pairs and two models. We see three  
patterns emerge.

\paragraph{Lightweight edits are largely ineffective for Qwen3-8B 
but not for Ministral-3-8B.}
For Qwen3-8B, \textbf{hedging} and \textbf{removal} resolve fewer 
than 7\% of detected issues across all language pairs. This suggests that Qwen3-8B 
does not condition strongly on individual trace steps when generating under replay. On the other hand, under our replay setup, Ministral is more sensitive to trace edits; the same lightweight edits resolve 41--60\% of issues.

\paragraph{Re-reasoning resolves issues but degrades translation quality.}
\textbf{Re-reasoning after removal} achieves resolution rates of 
50--63\% for Qwen3-8B, second only to hindsight. However, this 
comes at a cost: COMET scores drop substantially (up to $-0.16$ 
for en$\to$ja), so while the specific detected error might disappear, the regenerated trace suffix introduces new 
problems that degrade translation quality. The pattern is consistent across both models and most language pairs.

\paragraph{Hindsight is the strongest intervention.}
\textbf{Hindsight}, or replaying with a new trace generated mapping source to the reference, achieves the highest resolution rates  (72--84\% for Qwen3-8B, 83--88\% for Ministral-3-8B) and the 
most consistently positive COMET deltas when compared to the other intervention types. 

% \paragraph{Implications.}
% \cb{Stub for human evaluation results; if there is a disagreement between what people think are errors and what LLMs think are errors, what does that mean?}
% Taken together, these results reveal the unexpected differences between \emph{issue resolution} and \emph{translation quality}. 
% Interventions that resolve the most issues (re-reasoning, hindsight) do so by discarding and replacing large portions of the original 
% trace. Interventions that preserve the trace (hedging, removal) maintain translation quality but fail to propagate the intended corrections through to the output.  Additionally, the differences between what we see in Qwen3 and Ministral indicate that trace sensitivity is not an inherent property of reasoning-augmented translation but rather a model-specific characteristic, with implications for whether trace-level auditing can serve as a practical tool for translation quality improvement.

%% file: sections/conclusion.tex
\section{Discussion \& Conclusion}
\paragraph{Frequency of errors (RQ1)}
Our human validation reveals different story between Urdu and Spanish. For English$\to$Urdu, annotators confirm 93.3\% of the translations as not being faithful semantically to the source (\autoref{tab:human_validation_summary}), and 93.1\% of reasoning errors being valid errors. Here, the errors the LLM-as-a-judge detects are real, and  annotators understand why they are errors with high confidence, and the translations themselves are poor.

For English$\to$Spanish, the story is a bit different. Annotators confirm only 50\% of the detected errors, and rate up to 80\% of the translations as correct. We interpret this setting involving LLM-as-a-judge's pedantry, flagging reasoning steps with imprecise but harmless grammatical rules, objecting to reasonable interpretations of ambiguous phrases, or potentially diving into ungrounded stylistic preferences as errors. Notably, annotators are more confident when rejecting detected errors (97.3\%), than when confirming them (88.9\%), suggesting that non-errors are clearly non-errors to them.

Additionally, we observe that among reasoning errors that annotators confirm are errors, 84.0\% are reflected in the Spanish output versus only 41.1\% in the Urdu output. This suggests that the LLM-as-a-judge's standards are high enough that only  consequential errors tend to propagate. In Urdu, many confirmed errors exist in the reasoning, but are either partially corrected downstream, or may manifest as different surface errors in the output, making the mapping between reasoning step to output harder to track.

\paragraph{Implications on reasoning error detection}
Where models qualitatively produce worse translations, like in English$\to$Urdu, the LLM-as-a-judge reliably identifies genuine reasoning errors that a human auditor would agree with. For translation directions like in English$\to$Spanish, where the model translates more effectively, the same judge is overly pedantic -- roughly half of the detected errors do not align with what expert humans would consider reasonable reasoning errors. This means the signal is noisy, and any downstream use of the judge's output (for example, for filtering training data, or self-correcting potential errors), should account for this in practice.

\paragraph{Mitigation of reasoning errors (RQ2)}
The intervention results (\S\ref{subsec:intervention-results}) 
show that targeted trace edits guided by detected errors rarely improve translation quality. Hedging and removal resolve fewer than 7\% of issues across all high-resource pairs and under 1\% for English$\to$Urdu---the model produces nearly 
identical translations no matter what happens to the error. Only interventions 
that discard or replace large portions of the trace (re-reasoning, hindsight, oracle) achieve more substantial resolution  rates, but with slight degradation of COMET. The human evaluation helps explain this pattern, and the explanation differs by language pair. For English$\to$Spanish, where annotators reject roughly 
half the judge's flags, the failure of lightweight interventions is unsurprising: you cannot improve a translation by correcting an error that was 
not consequential in the first place. For English$\to$Urdu, however, the errors are more genuinely reasonable and the translations are poor, yet 
targeted edits still fail. Even when we know the reasoning is wrong and we know it affects the output, precisely repairing the reasoning step does not fix the translation in this setting. 

\paragraph{Towards better reasoning for translation.}
Our findings largely agree with prior work suggesting 
that native reasoning for MT does not work out of the 
box. We take a deeper dive and explore whether there 
are errors that might emerge from inconsistencies in the 
reasoning itself, finding that their properties differ across language pairs. These results suggest that reasoning for translation should not be free-form, deliberately exhaustive and ``over-reasoned'', or decoupled from the output. 
Instead, we argue that reasoning should be \emph{selective} (triggered only when warranted), \emph{faithful} (it should be enforced that trace decisions can be reflected in the translation output), and \emph{structured} (organized around verifiable sub-tasks rather than open-ended 
monologue). Our detection framework and subsequent error analysis offer a starting point for measuring progress along these axes.

%% file: figures/human_eval.tex
\begin{figure*}[t]
\centering
\footnotesize
\begin{tikzpicture}[
  box/.style={
    draw,
    rounded corners=7pt,
    fill=gray!10,
    align=left,
    inner sep=5pt,
    text width=7.25cm
  },
  annot/.style={
    draw,
    rounded corners=7pt,
    fill=blue!5,
    align=left,
    inner sep=5pt,
    text width=7.25cm
  },
  record/.style={
    draw,
    rounded corners=7pt,
    fill=green!5,
    align=left,
    inner sep=5pt,
    text width=7.25cm
  },
  arrow/.style={-Latex, thick}
]

% ---------------- Panel A ----------------
\node[font=\bfseries, align=center] (titleA) at (-4.4,5.8) {(A) Translation correctness};

\node[box, anchor=north west] (inputA) at (-8.5,5.4) {%
\textbf{Shown to annotator}\\
\textbf{Source sentence (English)}\\
\emph{[source sentence]}\\[0.35em]
\textbf{Translation into [language]}\\
\emph{[model translation]}
};

\node[annot, below=4mm of inputA] (annotA) {%
\textbf{Human bilingual annotator}\\
\textbf{Q1.} Does the translation preserve the important meaning of the source?\\
YES = no major meaning errors\\
NO = at least one meaning error that could mislead a reader\\
UNSURE = cannot confidently judge\\[0.35em]
\textbf{If Q1 = NO:} highlight the \emph{shortest minimal span} in the source that is mistranslated, and the \emph{shortest minimal span} in the translation that contains the error.
};

\node[record, below=4mm of annotA] (recordA) {%
\textbf{Stored annotation record}\\
translation\_correct $\in \{$YES, NO, UNSURE$\}$\\
source\_error\_span (if NO)\\
translation\_error\_span (if NO)
};

\draw[arrow] (inputA.south) -- (annotA.north);
\draw[arrow] (annotA.south) -- (recordA.north);

% ---------------- Panel B ----------------
\node[font=\bfseries, align=center] (titleB) at (4.4,5.8) {(B) Reasoning error validation};

\node[box, anchor=north west] (inputB) at (0.2,5.4) {%
\textbf{Shown to annotator}\\
\textbf{Automated detector flags a candidate reasoning span}\\[0.35em]
\textbf{Source sentence (English)}\\
\emph{[source sentence]}\\[0.35em]
\textbf{Translation into [language]}\\
\emph{[model translation]}\\[0.35em]
\textbf{Trace excerpt}\\
\emph{... [FLAGGED SPAN] ...}
};

\node[annot, below=4mm of inputB] (annotB) {%
\textbf{Human bilingual annotator}\\
\textbf{Q1.} Is the highlighted reasoning span actually an error?\\
YES / NO / BORDERLINE\\[0.3em]
\textbf{Q2.} How confident are you in the judgment above?\\
Confident / Somewhat confident / Not confident \\[0.3em]
\textbf{Q3.} If Q1 = YES or BORDERLINE, is this reasoning step reflected in the final translation?\\
YES / NO / UNSURE\\[0.3em]
\textbf{Q4.} Which of the following categories apply to this highlighted reasoning step? (multi-select): \\
Source misinterpretation; Internal contradiction; No issue; Other / Unsure
};

\node[record, below=4mm of annotB] (recordB) {%
\textbf{Stored annotation record (one per flagged span)}\\
q1\_is\_error $\in \{$YES, NO, BORDERLINE$\}$\\
q2\_reflected\_translation $\in \{$YES, NO, UNSURE$\}$\\
q3\_labels (multi-select)
};

\draw[arrow] (inputB.south) -- (annotB.north);
\draw[arrow] (annotB.south) -- (recordB.north);

\end{tikzpicture}
\caption{Bilingual human annotation protocol used to validate final translation correctness and detected reasoning errors. Part (A) is applied once per translation. Part (B) is applied once per detected reasoning error. Q3 is multi-select. For cost reasons, we run this protocol on English $\rightarrow$ Spanish and English $\rightarrow$ Urdu.}
\label{fig:bilingual_annotation_protocol}
\end{figure*}
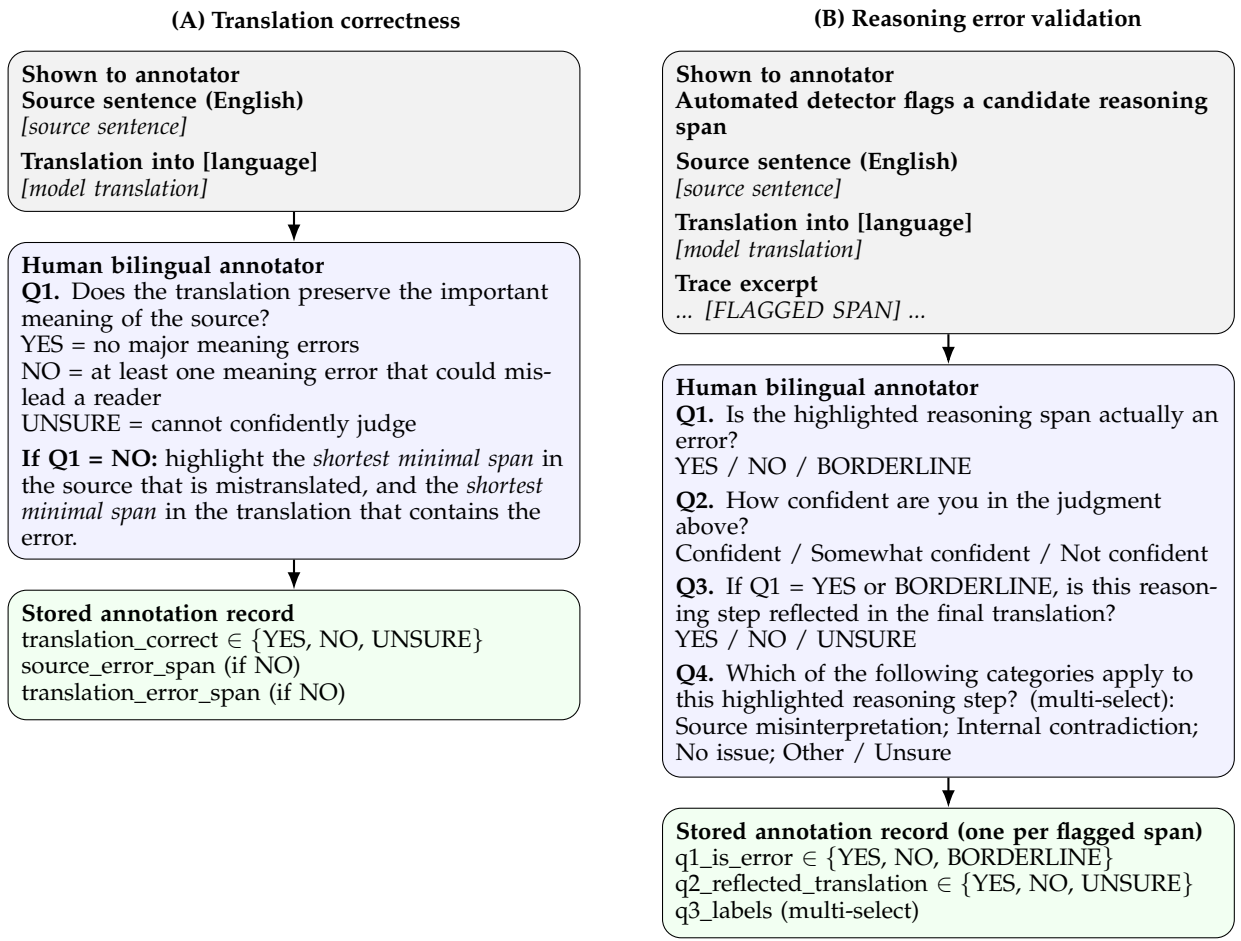